\documentclass{article}
\usepackage{arxiv_style,times}

\usepackage[utf8]{inputenc} % allow utf-8 input
\usepackage[T1]{fontenc}    % use 8-bit T1 fonts
\makeatletter
\def\zpl@scaled{s*[.97]}%
\InputIfFileExists{t1zpllf.fd}{}{}
\@ifundefined{T1/zpllf/m/n}{%
  \DeclareFontFamilySubstitution{T1}{zpllf}{zplLF}%
}{}
\makeatother
\usepackage{hyperref} % default link boxes, including green citation boxes
\usepackage{url}            % simple URL typesetting
\usepackage{booktabs}       % professional-quality tables
\usepackage{longtable}      % expression repertoire across appendix pages
\usepackage{amsfonts}       % blackboard math symbols
\usepackage{amssymb}
\usepackage{amsmath}
\usepackage{BOONDOX-cal}
\usepackage{nicefrac}       % compact symbols for 1/2, etc.
\usepackage{microtype}      % microtypography
\usepackage{xcolor}         % colors
\usepackage{colortbl}       % native colored cells in qualitative tables
\usepackage{graphicx}
\usepackage{multirow}
\usepackage{placeins}
\usepackage{wrapfig}        % wrapped dataset-statistics table
\usepackage{array}
\usepackage{tikz}
\usepackage{fontawesome5}
\usepackage[ruled,vlined,linesnumbered]{algorithm2e}
\usepackage[capitalize,nameinlink]{cleveref}
\crefname{table}{Tab.}{Tabs.}
\crefname{figure}{Fig.}{Figs.}
\crefname{section}{Sec.}{Secs.}
\crefname{subsection}{Sec.}{Secs.}
\crefname{equation}{Eq.}{Eqs.}
\crefname{algocf}{Alg.}{Algs.}

\usetikzlibrary{arrows.meta,positioning,decorations.pathmorphing,decorations.pathreplacing,calc,shapes.geometric,backgrounds,shadows}

\pgfdeclarelayer{background}
\pgfsetlayers{background,main}

\definecolor{junglegreen}{rgb}{0.113, 0.639, 0.5}
\DeclareRobustCommand{\BrainTwoFace}{{\normalfont\itshape Brain2Face}}

\newcolumntype{R}[1]{>{\raggedleft\arraybackslash}p{#1}}

\renewcommand{\tablename}{Tab.}

\title{Brain2Face: A Study on the Relation of EEG Signals and Spatio-temporal Head Avatars}
\author{%
Camilo Andrés Martínez Martínez\textsuperscript{1}
\And Olaf Dünkel\textsuperscript{1}
\And Kartik Teotia\textsuperscript{1}
\AND Maurice Rekrut\textsuperscript{2}
\And Christian Theobalt\textsuperscript{1}
\And Marc Habermann\textsuperscript{1}
}
\affiliations{%
\centering
\textsuperscript{1}Max Planck Institute for Informatics, Saarland Informatics Campus\\
\textsuperscript{2}German Research Center for Artificial Intelligence (DFKI)
}
\date{}
\hypersetup{%
    pdftitle={Brain2Face: A Study on the Relation of EEG Signals and Spatio-temporal Head Avatars},
    pdfauthor={Camilo Andrés Martínez Martínez, Olaf Dünkel, Kartik Teotia, Maurice Rekrut, Christian Theobalt, Marc Habermann}
}

\begin{document}

\maketitle

\begin{abstract}
%%%%
%
%%%%%%%%%%%%%%%%%%%%%%%%%
%
Driving digital head avatars is classically done by hand-crafted artist design or by image-based facial performance capture. In this work, we ask whether spatio-temporal tracking of the humans's head geometry and appearance can be performed solely from brain signals, i.e., Electroencephalography (EEG). Since prior work mostly focused on activity and emotion recognition, this task is relatively unexplored without even having publicly available high-quality datasets and benchmarks. Thus, we contribute a dataset featuring paired EEG and human head geometry as well as appearance annotations. We synchronize 64-channel EEG with per-frame expression latent codes of a personalized Gaussian head avatar. Second, we introduce a baseline to the task of expression recovery of a head avatar from EEG signals. To this end, we propose a dedicated EEG-to-expression model that takes a temporal sequence of EEG signals as input and regresses the latent expression parameters of the 3D Gaussian head avatar. Lastly, we validate our capture and algorithmic design choices through a benchmark protocol that we introduce to this task. We believe this work paves the way towards better understanding of the relation between EEG signals and facial surface as well as appearance changes. 
Project page: \url{https://vcai.mpi-inf.mpg.de/projects/Brain2Face/}.
%%%%%%%%%%%%%%%%%%%%%%%%%
%

%%%%
\end{abstract}

%%%%
        %
%%%%%%%%%%%%%%%%%%%%%%%%%%%%%%%%%
%
\section{Introduction} \label{sec:intro}

Digital human avatars play a central role in virtual reality, telepresence, and human-computer interaction, where the ability to faithfully convey facial expressions is crucial for realism and communication.
Existing avatar control paradigms typically rely on manual digital artist design or on visual observations, such as monocular or multi-view images, depth sensors, or facial markers, which inherently require a clear line of sight and explicit user engagement.
In contrast, brain-computer interfaces (BCIs) offer a fundamentally different interaction modality by enabling direct control from neural signals that do not suffer from such limitations:
Leveraging brain signals to drive expressive facial avatars would enable hands-free, camera-free avatar control, opening new possibilities for immersive communication, accessibility, and assistive technologies.
Among the devices that measure brain activities, electroencephalography (EEG) stands out due to its non-invasive nature and practical usage.
However, current EEG technology still poses fundamental challenges for actual deployment, e.g., due to its setup and noisy signals and it remains an open question to what extent EEG signals are already usable for such deployments.
% ------------------
%
\par
On the other hand, image-based facial performance capture has been studied extensively, from multi-view reconstruction to learned facial models and photorealistic avatars~\citep{beeler2010multiview,danecek2022emoca,qian2024gaussianavatars}.
Meanwhile, EEG-based interfaces have primarily focused on discrete prediction tasks, including affective state recognition~\citep{DEAPdataset2012,MAHNOBdataset2012} and clinical pathology detection~\citep{GEMEIN2020117021,HBM:HBM23730}, rather than continuous facial motion.
Related synthesis approaches generate facial emotion images~\citep{esmaeili2024generating}, EEG-conditioned talking faces~\citep{park2024towards}, or intentional facial expressions through brain decoding and generative image synthesis~\citep{shimizu2025reviving}.
In contrast, we predict per-frame expression codes of a 3D head avatar rather than images.
Furthermore, multi-view capture systems~\citep{beeler2010multiview} and expressive photorealistic head models~\citep{qian2024gaussianavatars,teotia2025audiodrivenuniversalgaussianhead} provide detailed geometry and appearance representations for facial animation.
Studies on EEG-driven photorealistic avatars, however, remain limited, e.g., Mind-to-Face~\citep{xiong2025mindtoface}.
Still, the capture strategy focuses on emotion-stimulating videos and does not provide a fixed set of recorded facial expression, limiting systematic analysis.
Further, captured modalities remain limited, i.e., limited number of EEG channels and views, and data is not publicly available for research.
%
%
% ------------------
%
\par
\begin{figure*}[t]
\begingroup
\centering
\makebox[\textwidth][c]{%
\begin{minipage}{\textwidth}
\centering
\includegraphics[width=\linewidth]{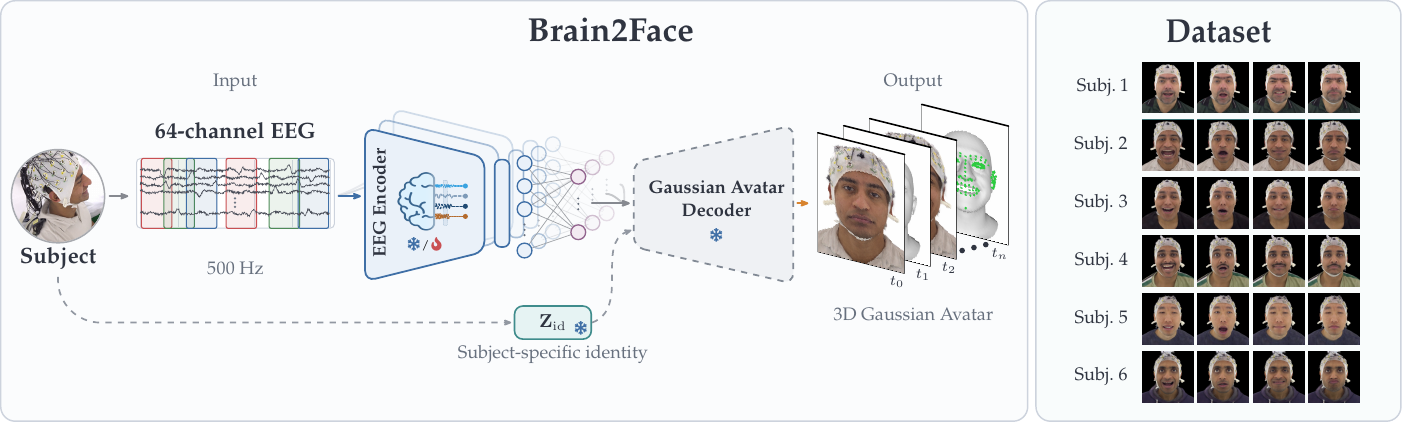}
\caption[\BrainTwoFace{} EEG-to-Face task and fitted avatar targets]{%
\textbf{Brain2Face.}
(\emph{Left}) A 64-channel EEG recording is encoded into a pre-trained latent expression of a Gaussian avatar prior \citep{teotia2025audiodrivenuniversalgaussianhead}.
Then, the frozen decoder of this prior takes the predicted expression and a subject-specific identity latent and regresses Gaussian splat parameters and geometry.
We note that acquiring the identity latent requires a personalization step.
(\emph{Right}) Fitted targets for six participants (rows) and four expressions (columns) of our newly acquired dataset.
}
\label{fig:teaser}
\end{minipage}%
}%
\endgroup
\end{figure*}

To address these limitations, we introduce a dataset of paired EEG and spatio-temporally coherent and fine-grained facial performance capture data, including geometry and appearance.
We captured multiple subjects in a multi-view capture setup with up to 33 cameras, paired with 64-channel EEG signals.
To recover high-quality face data, we fit a state-of-the-art 3D Gaussian avatar to the data and, thus, encode expressions into low dimensional latent codes.
An EEG encoder and a decoder map temporal EEG signals to these expression latents.
This allows us to decode the EEG driven expression into 3D Gaussians and geometry that can be rendered to image space.
For training, supervision is applied in latent and geometry space.
We refer to Fig.~\ref{fig:teaser} for an overview.
\par 
Since this is the first study of its kind, we introduce an evaluation protocol for this task and a first baseline as well as ablations of various EEG foundation models~\citep{ouahidi2025reve,guetschel2024sjepa}, the influence of different prediction heads, and the effect of masking different brain areas.
We believe that our benchmark can serve as a starting point for future works to study the relations between EEG data and facial expressions.
%
% ------------------
%
\par
To summarize, this work makes several key contributions:
\begin{itemize}
\item First, we present a dataset that provides EEG signals paired with high-resolution facial performance capture data as well as dense annotations that are encoded as latent expression parameters.
\item Second, we propose a model for predicting facial expressions from EEG signals, which allows driving a Gaussian head avatar.
\item Third, we ablate several baselines and design factors on our newly introduced benchmark.
\end{itemize}
%
%%%%%%%%%%%%%%%%%%%%%%%%%%%%%%%%%
%

%%%%

%%%%
% Brain2Face teaser figure.
%%%%%%%%%%%%%%%%%%%%%%%%%%%%%%%%%%%%%%
%
% Brain icon adapted from the braindecode logo: https://braindecode.org/.
%
%%%%%%%%%%%%%%%%%%%%%%%%%%%%%%%%%%%%%%
%

%%%%

%%%%
%
% ================================
%
\section{Related Work} \label{sec:rw}
%
% ------------------
%
\par
\paragraph{EEG-based Brain-Computer Interfaces.}
Electroencephalography (EEG) captures the summation of postsynaptic potentials from cortical pyramidal neurons, providing high-temporal-resolution voltage fluctuations measured at the scalp \cite{gkintoni2025mapping}.
Since its discovery by Hans Berger in the 1920s, EEG has evolved from a clinical device into a versatile tool for decoding neural activity in both constrained laboratory and everyday environments \cite{gkintoni2025mapping,beyrouthy2024review}.
The majority of EEG-based brain-computer interface (BCI) research has focused on \emph{classification} tasks, mapping neural signals to a small, fixed set of categories.
A prominent line of work concerns affective state recognition, leveraging multi-modal datasets such as DEAP~\cite{DEAPdataset2012} and MAHNOB~\citep{MAHNOBdataset2012} to classify valence or arousal labels.
Recent methods combine EEG with video and audio streams to improve emotion classification through multimodal fusion \citep{wang2025milmerframeworkmultipleinstance}.
In the clinical domain, deep convolutional and temporal convolutional networks have demonstrated strong performance for discrete EEG pathology detection \cite{GEMEIN2020117021,HBM:HBM23730}, and the Braindecode toolbox \cite{HBM:HBM23730} has made such architectures widely accessible.
More recently, large-scale EEG foundation models have been proposed to learn general-purpose representations transferable across subjects and tasks~\citep{ouahidi2025reve}.
A smaller body of work has explored \emph{continuous regression} from EEG, targeting low-dimensional motor outputs.
\cite{hosseini2022handmovementeeg} decode two-dimensional hand-movement trajectories from 63-channel EEG using phase-based connectivity features and linear regression, while \cite{lu2026electroencephalography} extend continuous decoding to polar-coordinate hand-motion angles.
Bridging neural signals and facial motion, \cite{shimizu2025reviving} interface brain decoding with generative image synthesis to restore intentional facial expressions for ALS patients, and \cite{park2024towards} generate lip-synchronised talking faces conditioned on EEG for dynamic communication.
\cite{esmaeili2024generating} synthesize discrete facial emotion images from EEG using conditional adversarial networks.
However, none of these works has investigated learning and studying a continuous mapping from EEG signals to facial geometry and appearance parameterized as a latent expression space of a Gaussian head avatar prior.
%
% ------------------
%
\par
\paragraph{Facial Performance Capture.}
Recovering fine-grained facial geometry and appearance from image streams has been studied extensively.
Monocular methods~\citep{feng2021deca, danecek2022emoca} fit parametric face models~\citep{blanz1999morphable, li2017flame} to single RGB images or video, leveraging shape, expression, and appearance priors to regress 3D morphable model coefficients or neural blendshape parameters.
Multi-view systems~\citep{beeler2010multiview} improve reconstruction fidelity by triangulating correspondences across calibrated cameras, enabling denser and more accurate surface recovery.
More recent approaches build implicit neural or Gaussian representations directly from video footage, capturing subject-specific geometry and view-dependent appearance with high visual fidelity~\citep{gafni2021nerface, zheng2022imavatar, qian2024gaussianavatars}.
While egocentric setups using head-mounted displays~\citep{zhang2024refa} have relaxed the requirement for external camera rigs, all of these paradigms share a common dependency: they require a clear line of sight to the subject's face and operate under visual sensing assumptions.
EEG needs no line of sight, but its setup and noisy, artifact-prone signals~\citep{goncharova2003emg} still hinder deployment.
Moreover, image-based methods cannot be trivially adapted to EEG inputs, since EEG signals have entirely different statistical properties, spatial structure, and signal-to-noise characteristics compared to pixel data, necessitating dedicated architectures and datasets.
%
% ------------------
%
\par
\paragraph{Photoreal Head Avatars.}
Recent years have seen remarkable advances in the fidelity of animatable head avatars.
Neural radiance field (NeRF) based approaches~\citep{mildenhall2020nerf, gafni2021nerface} and, more recently, 3D Gaussian splatting representations~\citep{kerbl2023gaussian, qian2024gaussianavatars, xu2024gaussianhead} have enabled photorealistic rendering of dynamic faces at interactive frame rates.
A range of methods construct person-specific neural avatars conditioned on expression codes derived from 3DMM fits~\citep{gafni2021nerface, zheng2022imavatar} or tracked blendshape rigs~\citep{qian2024gaussianavatars}, achieving impressive reconstruction quality.
Universal avatar priors that generalize across identities have been proposed to disentangle subject appearance from expression control, allowing a single model to animate novel subjects given only a short capture~\citep{teotia2025audiodrivenuniversalgaussianhead}.
In this work, we adopt such a universal Gaussian head avatar as our rendering backbone and treat its learned expression latent space as the regression target for our EEG encoder.
Our goal is to study EEG-to-expression regression using a learned avatar latent space and to compare EEG representations under controlled evaluation protocols.
%
%%%%%%%%%%%%%%%%%%%%%%%%%%%%%%%%%
%
%
% ------------------
%
\par
The most closely related work is Mind-to-Face~\citep{xiong2025mindtoface}, which also reconstructs photorealistic, dynamic 3D facial expressions from EEG.
Their system records synchronized 16-channel EEG and multi-view video during emotion-eliciting stimuli videos, maps EEG through a CNN-Transformer encoder to dense 3D position maps, and renders the resulting geometry with a Gaussian avatar pipeline.
Their study evaluates personalized mappings for two subjects.
Brain2Face instead maps 64-channel EEG to the 256-dimensional expression codes of a personalized universal Gaussian head avatar and studies six subjects performing a predefined set of 75 expression types across two sessions.
This recording allows systematic analyses, e.g., the comparison of the effectiveness of various EEG foundation models, the generalizability of trained EEG-to-face models on seen and unseen expressions, or studies of sensitivity with respect to masked electrodes.
Therefore, our dataset offers annotations previously not available, enabling a a wide range of potential analyses to better understand the relation between EEG signals and facial expression.

%%%%
%
% ================================
%
\section{Brain2Face Dataset} \label{sec:dataset}
%
%
%%%%
%
%%%%%%%%%%%%%%%%%%%%%%%%%%%%%%%%%%%%%%
%
%
\begin{figure}[t]
    \centering
    \resizebox{\linewidth}{!}{%
    \begin{tikzpicture}[
            soft shadow/.style={
                    preaction={fill=black, opacity=0.08,
                            transform canvas={xshift=0.4mm, yshift=-0.4mm}}
                },
            imgnode/.style={inner sep=0pt, draw=gray!50, line width=0.5pt,
                    soft shadow},
            rowlabel/.style={font=\small, rotate=90, anchor=center},
        ]
        \def\imgW{2.25}
        \def\imgH{2.25}
        \def\imgGap{0.10}
        \def\xStart{1.45}
        \def\xMax{13.10}
        \def\yOne{0.00}
        \def\yTwo{-2.85}
        \def\yThree{-5.70}
        \node[rowlabel] at (\xStart - 0.40, \yOne)
        {Expressions};
        \node[rowlabel] at (\xStart - 0.40, \yTwo)
        {Views};
        \node[rowlabel] at (\xStart - 0.40, \yThree)
        {Subjects};
        {
                \pgfmathsetmacro{\cx}{\xStart + 0*(\imgW+\imgGap) + \imgW/2}
                \node[imgnode] at (\cx, \yOne)
                {\includegraphics[width=\imgW cm, height=\imgH cm]%
                    {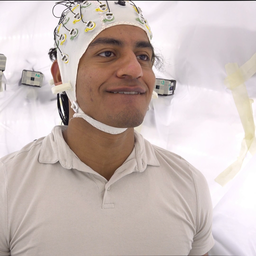}};
        }
        {
                \pgfmathsetmacro{\cx}{\xStart + 1*(\imgW+\imgGap) + \imgW/2}
                \node[imgnode] at (\cx, \yOne)
                {\includegraphics[width=\imgW cm, height=\imgH cm]%
                    {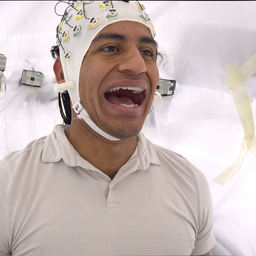}};
        }
        {
                \pgfmathsetmacro{\cx}{\xStart + 2*(\imgW+\imgGap) + \imgW/2}
                \node[imgnode] at (\cx, \yOne)
                {\includegraphics[width=\imgW cm, height=\imgH cm]%
                    {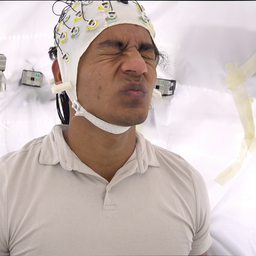}};
        }
        {
                \pgfmathsetmacro{\cx}{\xStart + 3*(\imgW+\imgGap) + \imgW/2}
                \node[imgnode] at (\cx, \yOne)
                {\includegraphics[width=\imgW cm, height=\imgH cm]%
                    {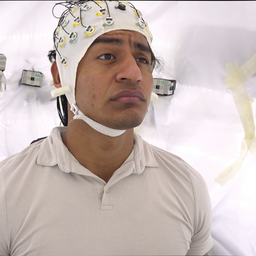}};
        }
        {
                \pgfmathsetmacro{\cx}{\xStart + 4*(\imgW+\imgGap) + \imgW/2}
                \node[imgnode] at (\cx, \yOne)
                {\includegraphics[width=\imgW cm, height=\imgH cm]%
                    {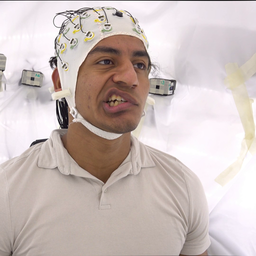}};
        }
        {
                \pgfmathsetmacro{\cx}{\xStart + 0*(\imgW+\imgGap) + \imgW/2}
                \node[imgnode] at (\cx, \yTwo)
                {\includegraphics[width=\imgW cm, height=\imgH cm]%
                    {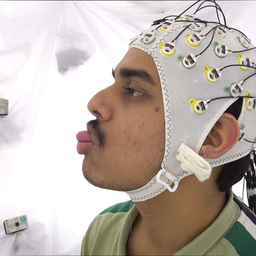}};
        }
        {
                \pgfmathsetmacro{\cx}{\xStart + 1*(\imgW+\imgGap) + \imgW/2}
                \node[imgnode] at (\cx, \yTwo)
                {\includegraphics[width=\imgW cm, height=\imgH cm]%
                    {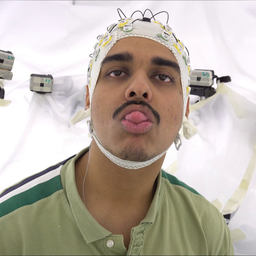}};
        }
        {
                \pgfmathsetmacro{\cx}{\xStart + 2*(\imgW+\imgGap) + \imgW/2}
                \node[imgnode] at (\cx, \yTwo)
                {\includegraphics[width=\imgW cm, height=\imgH cm]%
                    {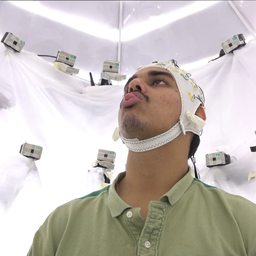}};
        }
        {
                \pgfmathsetmacro{\cx}{\xStart + 3*(\imgW+\imgGap) + \imgW/2}
                \node[imgnode] at (\cx, \yTwo)
                {\includegraphics[width=\imgW cm, height=\imgH cm]%
                    {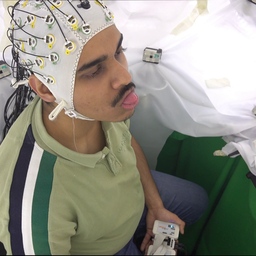}};
        }
        {
                \pgfmathsetmacro{\cx}{\xStart + 4*(\imgW+\imgGap) + \imgW/2}
                \node[imgnode] at (\cx, \yTwo)
                {\includegraphics[width=\imgW cm, height=\imgH cm]%
                    {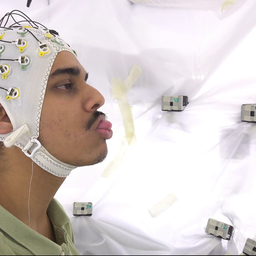}};
        }
        {
                \pgfmathsetmacro{\cx}{\xStart + 0*(\imgW+\imgGap) + \imgW/2}
                \node[imgnode] at (\cx, \yThree)
                {\includegraphics[width=\imgW cm, height=\imgH cm]%
                    {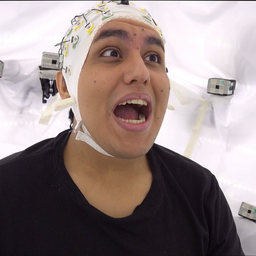}};
        }
        {
                \pgfmathsetmacro{\cx}{\xStart + 1*(\imgW+\imgGap) + \imgW/2}
                \node[imgnode] at (\cx, \yThree)
                {\includegraphics[width=\imgW cm, height=\imgH cm]%
                    {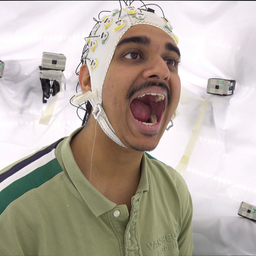}};
        }
        {
                \pgfmathsetmacro{\cx}{\xStart + 2*(\imgW+\imgGap) + \imgW/2}
                \node[imgnode] at (\cx, \yThree)
                {\includegraphics[width=\imgW cm, height=\imgH cm]%
                    {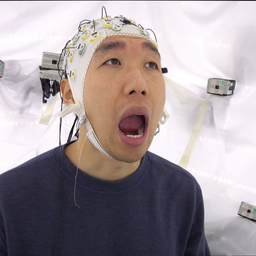}};
        }
        {
                \pgfmathsetmacro{\cx}{\xStart + 3*(\imgW+\imgGap) + \imgW/2}
                \node[imgnode] at (\cx, \yThree)
                {\includegraphics[width=\imgW cm, height=\imgH cm]%
                    {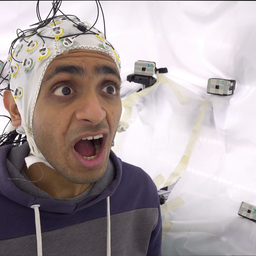}};
        }
        {
                \pgfmathsetmacro{\cx}{\xStart + 4*(\imgW+\imgGap) + \imgW/2}
                \node[imgnode] at (\cx, \yThree)
                {\includegraphics[width=\imgW cm, height=\imgH cm]%
                    {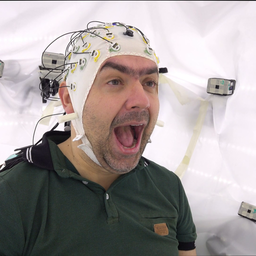}};
        }
        %
        % ---- Background card ----
        \begin{pgfonlayer}{background}
            \fill[blue!2, rounded corners=6pt]
            (-0.10, 1.70) rectangle (\xMax + 0.10, -7.30);
        \end{pgfonlayer}
    \end{tikzpicture}%
    }
    \caption{
        Overview of the \textbf{Brain2Face} dataset.
        Our dataset comprises paired EEG and multi-view video data covering multiple facial expressions, views, and subjects.
    }
    \label{fig:dataset}
\end{figure}
%
%%%%%%%%%%%%%%%%%%%%%%%%%%%%%%%%%%%%%%
%

%%%%
%
%
% ------------------
%
\par
The Brain2Face dataset pairs 64-channel EEG with multi-view recordings of controlled facial expressions.
Personalized Gaussian head avatars~\citep{teotia2025audiodrivenuniversalgaussianhead} provide continuous expression targets disentangled from subject identity.
In this section, we describe the capture setup and avatar fitting.
Tab.~\ref{tab:dataset-statistics} summarizes the statistics of the collected dataset and Fig.~\ref{fig:dataset} shows qualitative examples.
%
% --------------------------------
%
\begin{wraptable}{r}{0.4\textwidth}
    \vspace{-10pt}
    \centering
    \caption{\textbf{Brain2Face dataset statistics.} Paired EEG and multi-view face recordings across all subjects and sessions.}
    \label{tab:dataset-statistics}
    \begingroup
    \footnotesize
    \setlength{\tabcolsep}{3pt}
    \renewcommand{\arraystretch}{1.0}
    \begin{tabular}{@{}lr@{}}
        \toprule
        Property & Value \\
        \midrule
        Subjects & 6 \\
        Sessions per subject & 2 \\
        Total capture time & 4~h 15~min \\
        Expression types per session & 75 \\
        %Paired segments (5~s each) & 891 \\
        Expression data & 74~min 15~s \\
        \addlinespace[3pt]
        EEG channels & 64 \\
        EEG sampling rate  & 500~Hz \\
        Cameras per session & 26--33 \\
        Video frame rate & 29.97~fps\\
        \bottomrule
    \end{tabular}
    \endgroup
    \vspace{-\intextsep}
    \vspace{-30pt}
\end{wraptable}

\subsection{Capture Setup}
%
% ------------------
%
\par
\paragraph{Face Capturing.}
Subjects are recorded by a multi-view setup of 26 to 33 cameras recording video at $f_v=29.97$~fps.
The multi-view images are used to fit a personalized Gaussian head avatar~\citep{teotia2025audiodrivenuniversalgaussianhead} to each subject (Sec.~\ref{subsec:avatar_fitting}), yielding per-frame expression codes and a fixed identity code.
The fitted expression codes provide continuous regression targets. Decoding them yields the meshes, 3D landmarks, and avatar renders used for supervision and evaluation (Sec.~\ref{subsec:avatar_fitting}).
%
% ------------------
%
\par
\paragraph{EEG Capturing.}
Electroencephalography (EEG) measures scalp voltage fluctuations arising from summed postsynaptic activity with high temporal resolution~\citep{gkintoni2025mapping}.
We acquire 64-channel EEG at 500~Hz using a Brain Products LiveAmp device with an EasyCap and electrode montage according to the 10-10 system ~\citep{oostenveld2001five}, with conductive gel controlling electrode impedance. Impedance was measured before the experiment and kept below 15 k$\Omega$.
%
%
% ------------------
%
\par
\paragraph{Recording Sessions.}

Every subject follows a standardized experimental protocol with a fixed set of expressions (Secs.~\ref{app:dataset-splits} and~\ref{app:expression-repertoire} in the supplementary material).
A baseline EEG recording quantifies the subject's resting neural activity.
Then, the subject performs a set of pre-defined facial expressions in two sessions, following video demonstrations of each target expression, separated by a short break.
Each session starts and ends with three jaw clenches that serve as a synchronization signal (Sec.~\ref{subsec:sync} in the supplementary material).
%
%
%%%%
%%%%
%
% --------------------------------
%
\subsection{Gaussian Avatar Fitting} \label{subsec:avatar_fitting}
%
% ------------------
%
\par
We build on the Universal Head Avatar Prior (UHAP) introduced by UniGAHA~\citep{teotia2025audiodrivenuniversalgaussianhead}, which disentangles identity and expression for photorealistic head rendering using 3D Gaussian splatting~\citep{kerbl2023gaussian}.
A frozen UHAP image encoder maps each frontal video frame to an expression code $\mathbf{Z}_{\mathrm{exp}}^{t}\in\mathbb{R}^{256}$.
Fitting UHAP to each subject's multi-view recordings yields an identity code $\mathbf{Z}_{\mathrm{id}}\in\mathbb{R}^{512}$ and a personalized decoder.
The personalized decoder $\mathcal{D}_{\mathrm{UHAP}}$ maps the identity and expression codes to $N_g$ Gaussian primitives and a guide mesh, $(\{\mathbf{g}_{i}^{t}\}_{i=1}^{N_g},\mathbf{V}^{t})=\mathcal{D}_{\mathrm{UHAP}}(\mathbf{Z}_{\mathrm{id}},\mathbf{Z}_{\mathrm{exp}}^t)$.
A differentiable Gaussian-splatting renderer $R$ produces the image $\mathbf{I}^{t}=R(\{\mathbf{g}_{i}^{t}\}_{i=1}^{N_g})$, and a fixed vertex-selection operator $\Pi$ extracts the 3D landmarks $\mathbf{L}^{t}=\Pi(\mathbf{V}^{t})$.
Camera arguments are suppressed for readability; predicted and fitted images use the same camera.
The images $\mathbf{I}^{t}$ are fitted-avatar renders, while the captured multi-view images are used to fit the avatar.
The personalized decoder and identity code remain frozen during EEG training.
Sec.~\ref{app:avatar-fitting} in the supplementary material describes fitting, inversion, and the construction of the fitted geometry.
%
% ------------------
%
\par
%
%
% --------------------------------
%
%
% ------------------
%
%
%%%%
%%%%
%
%
%
% ================================
%
\section{Method} \label{sec:method}
%
%
%%%%
%%%%
%
% ------------------
%
\par
The objective of \emph{Brain2Face} is continuous control of a personalized 3D Gaussian head avatar from EEG signals.
Given a preprocessed EEG window $\mathbf{X}\in\mathbb{R}^{C\times N_e}$, a predictor $F$ estimates the corresponding expression-code sequence $\hat{\mathbf{Z}}_{\mathrm{exp}}=F(\mathbf{X})\in\mathbb{R}^{d\times N_z}$, whose target $\mathbf{Z}_{\mathrm{exp}}$ stacks the codes $\mathbf{Z}_{\mathrm{exp}}^{t}$ of the window's $N_z$ frames, where $C=64$, $d=256$, and $N_e$ and $N_z$ denote the numbers of EEG samples and target frames.
Each predicted code drives the frozen personalized avatar decoder defined in Sec.~\ref{subsec:avatar_fitting}.
Fig.~\ref{fig:overview} in the supplementary material gives an overview of the pipeline.
%

%
% --------------------------------
%
\subsection{EEG-to-Expression Encoder} \label{subsec:eegencoder}
%
% ------------------
%
\par
\paragraph{Data preprocessing.}
We follow the pretraining preprocessing of REVE~\citep[Sec.~3.1.1]{ouahidi2025reve}: resampling to 200~Hz, 0.5--99.5~Hz band-pass filtering, and clipping at $\pm15$ standard deviations per session.
Further, we employ a notch filter at 50~Hz to remove power line noise.
%
% ------------------
%
\par
\paragraph{Model architecture.}
We factor the predictor as $F=\mathcal{M}_{\phi}\circ\mathcal{E}_{\theta}$, where $\mathcal{E}_{\theta}$ encodes the EEG window and $\mathcal{M}_{\phi}$ maps its features to the expression-code sequence.
The encoder can be initialized from an EEG foundation model and either frozen or fine-tuned.
For the EEG encoder, we use the base variant of REVE~\citep{ouahidi2025reve}.
The decoder flattens the encoder's channel and patch dimensions, projects the features with a $1\times1$ convolution, and applies $L$ residual convolution blocks along the resulting channel-patch token axis.
A learned resampler and a framewise output projection produce $N_z$ expression codes with $d=256$ coordinates each.
We use $L=2$ and one-second windows ($N_e=200$, $N_z=30$) for the principal model and study the decoder depth $L$ in Sec.~\ref{sec:reve-initialization-depth}.
During training, an auxiliary head pools the intermediate residual-block features and maps them to a second expression-code sequence $\hat{\mathbf{Z}}_{\mathrm{exp}}^{\mathrm{aux}}$ of the same shape, which supervises the decoder's intermediate representations and is discarded at inference.
The foundation-model comparison instead uses a decoder with two fully connected hidden layers to probe each encoder's frozen features (Sec.~\ref{sec:foundation-model-selection}).
Sec.~\ref{subsec:implementation} provides further details.
%

% ------

%
% ------------------
%
\par
Replacing $\mathbf{Z}_{\mathrm{exp}}^t$ with $\hat{\mathbf{Z}}_{\mathrm{exp}}^t$ in the same decoder and renderer gives the predicted mesh $\hat{\mathbf{V}}^{t}$, landmarks $\hat{\mathbf{L}}^{t}$, and image $\hat{\mathbf{I}}^{t}$.
%
%
% --------------------------------
%
\subsection{Supervision} \label{subsec:supervision}
%
% ------------------
%
\par
We supervise the predicted expression latents and decoded geometry with the corresponding fitted-avatar targets
with the following training objective:
%%%%
\begin{equation}
    \begin{aligned}
    \mathcal{L}_{\mathrm{train}}={}
    \lambda_p\mathcal{L}_p(\hat{\mathbf{Z}}_{\mathrm{exp}},\mathbf{Z}_{\mathrm{exp}})
    +\lambda_v\mathcal{L}_v(\hat{\mathbf{V}},\mathbf{V})
    +\lambda_\ell\mathcal{L}_\ell(\hat{\mathbf{L}},\mathbf{L})
    +\lambda_{\mathrm{aux}}\mathcal{L}_{\mathrm{aux}}(\hat{\mathbf{Z}}_{\mathrm{exp}}^{\mathrm{aux}},\mathbf{Z}_{\mathrm{exp}}).
    \end{aligned}
    \label{eq:total_loss}
\end{equation}
%%%%
Each loss averages over the frames and coordinates of a training batch (Sec.~\ref{subsec:training_objective}).
The first three terms measure latent mean squared error, mesh mean absolute error, and the landmark Wing loss~\citep{feng2018wing}, which combines a logarithmic penalty on small landmark errors with an $\ell_1$ penalty on large ones.
The auxiliary term is the mean squared error between the target codes and the auxiliary sequence $\hat{\mathbf{Z}}_{\mathrm{exp}}^{\mathrm{aux}}$ predicted from intermediate decoder features. % (Sec.~\ref{subsec:eegencoder}).
It regularizes those features by requiring them to already decode the full expression trajectory.
Geometry gradients pass through the frozen avatar decoder to the trainable predictor parameters.
Exact loss definitions and weights are given in Sec.~\ref{subsec:training_objective}, with optimization and inference details in the supplementary material.

% %
% %%%%
% %%%%
% %
% %
% %
% %
% %
%%%%
% %
% %
% %
% %

% %
% %%%%%%%%%%%%%%%%%%%%%%%%%%%%%%%%%
% %

% %

% %
% %
% %
% %
% %
%%%%
% %
% %
% %
%%%%
% %
% %
% %
%%%%

% %
% %%%%%%%%%%%%%%%%%%%%%%%%%%%%%%%%%
% %

% %
% %
%%%%
% %
% %
% %

% %
% %
% %
% %
% %

% %
% %
% %
%%%%
% %
% %
%%%%
% %
% %
% %
%%%%
% %
% %
% %

% %
% %
%%%%
% %
% %

% %
% %%%%%%%%%%%%%%%%%%%%%%%%%%%%%%%%%
% %

% %
% %
% %
%%%%
% %
% %
%%%%
% %
% %
% %
% %
%%%%
% %
% %
% 3D facial landmarks are extracted by a fixed vertex selection operator $\Pi$:
% %
%%%%
% %
% %
%%%%
% %

% %
% %%%%%%%%%%%%%%%%%%%%%%%%%%%%%%%%%
% %

% %
% %
%%%%
% %
% %
%%%%
% %
% %
% %
%%%%
% %
% %
%%%%
% %
% %
%%%%
% %
% %
% %
% %

% %
% %%%%%%%%%%%%%%%%%%%%%%%%%%%%%%%%%
% %

% %
% %
% %
% %
% %
% %
% %
% %%%%%%%%%%%%%%%%%%%%%%%%%%%%%%%%%
% %

%%%%
%
%%%%%%%%%%%%%%%%%%%%%%%%%%%%%%%%%
%
\FloatBarrier
%
% ================================
%
\section{Experiments} \label{sec:experiments}
%
% ------------------
%
\par
Next, we study the feasibility of predicting facial expressions from EEG signals.
We introduce the evaluation protocol (Sec.~\ref{sec:evaluation-protocol}), report our experimental findings (Sec.\ref{sec:foundation-model-selection}-\ref{sec:area}), and, finally, show qualitative results of the overall pipeline (Sec.~\ref{sec:qualitative-seen-unseen}).
%
% --------------------------------
%
\subsection{Evaluation Protocol}
\label{sec:evaluation-protocol}
%
% ------------------
%
\par
\paragraph{Metrics.}
To quantitatively evaluate the performance of the face predictions, we use the following evaluation metrics:
Latent MSE measures the mean squared error between predicted and target expression codes.
Mesh MAE measures the mean absolute coordinate error over the decoded mesh vertices.
Landmark Wing measures a robust coordinate-wise deviation of the selected facial landmarks~\citep{feng2018wing}.
Blendshape MAE measures the mean absolute error across 52 expression coefficients.
LPIPS~\citep{zhang2018lpips} measures perceptual distance between predicted and fitted-avatar ground-truth renders.
Sec.~\ref{subsec:training_objective} defines latent MSE, mesh MAE, and landmark Wing as $\mathcal{L}_p$, $\mathcal{L}_v$, and $\mathcal{L}_\ell$, evaluated on the test frames.
All metrics are averaged over all frames within each recording-expression held-out sequence.
% ------------------
%
\par
%
%
% ------------------
%
\par
\paragraph{Test splits.}
We train and evaluate a separate model for each subject using the cross-session split described in Sec.~\ref{app:dataset-splits}.
Both test settings contain sequences excluded from training.
For \textit{seen expressions} (S), the expression type occurs in training in the opposite recording session; for \textit{unseen expressions} (U), it is absent from training in both sessions.
Both sessions contribute training sequences, so this protocol does not evaluate transfer to a completely unseen recording session or participant.
%
%

%
% --------------------------------
%
\subsection{On the effectiveness of EEG foundation models for face decoding}
\label{sec:foundation-model-selection}
%
% ------------------
%
\par
We compare nine pretrained EEG encoders $\mathcal{E}_{\theta}$ by freezing their parameters and training an MLP decoder $\mathcal{M}_{\phi}$ with two fully connected hidden layers to predict expression-code trajectories.
Each predictor maps the full EEG input window to the corresponding target trajectory.
Tab.~\ref{tab:foundation-model-comparison} reports results on Subject-03 test sequences with seen expression types.
%
%%%%
\begin{table}[!t]
    \centering
    \caption{\textbf{Frozen EEG foundation model comparison.}
    Results on test sequences with seen expressions for frozen pretrained EEG encoders
    with a nonlinear MLP decoder, evaluated on Subject-03.
    REVE [scratch] trains the same encoder and decoder from random initialization,
    with the encoder unfrozen, with the same data and compute.
    }
    \label{tab:foundation-model-comparison}
    \vspace{3pt}

    \begingroup
    \footnotesize
    \setlength{\tabcolsep}{3pt}
    \renewcommand{\arraystretch}{1.08}
    \resizebox{\textwidth}{!}{%
    \begin{tabular}{lccccc}
        \toprule
        \textbf{Model}
            & \textbf{LPIPS} $\downarrow$
            & \textbf{Blendshape MAE} $\downarrow$
            & \textbf{Latent MSE} $\downarrow$
            & \textbf{Mesh MAE [mm]} $\downarrow$
            & \textbf{Lmk Wing [mm]} $\downarrow$ \\
        \midrule
        BENDR~\citep{kostas2021bendr} & 0.0607 & 0.0259 & 0.242 & 0.906 & 2.737 \\
        BIOT~\citep{yang2023biot} & 0.0671 & 0.0570 & 0.247 & 1.106 & 2.768 \\
        CBraMod~\citep{wang2025cbramod} & 0.0624 & 0.0266 & 0.253 & 0.914 & 2.791 \\
        CodeBrain~\citep{ma2026codebrain} & 0.0587 & 0.0249 & 0.231 & 0.748 & 2.563 \\
        EEGPT~\citep{wang2024eegpt} & 0.0659 & 0.0567 & 0.247 & 0.877 & 2.754 \\
        LaBraM~\citep{jiang2024labram} & 0.0817 & 0.0476 & 0.236 & 1.308 & 2.751 \\
        LUNA ~\citep{doner2025luna} & 0.0704 & 0.0558 & 0.254 & 0.990 & 2.883 \\
        Signal-JEPA~\citep{guetschel2024sjepa} & 0.0669 & 0.0566 & 0.247 & 1.003 & 2.742 \\
        REVE~\citep{ouahidi2025reve} & \textbf{0.0575} & \textbf{0.0238} & \textbf{0.209} & \textbf{0.725} & \textbf{2.499} \\
        REVE [scratch] & 0.0603 & 0.0258 & 0.254 & 1.713 & 2.737 \\
        \bottomrule
    \end{tabular}%
    }
    \endgroup
\end{table}

%%%%
%
% ------------------
%
\par
Among the frozen pretrained encoders, REVE achieves the lowest error on all five reported metrics.
We hypothesize that this stems mainly from pretraining: REVE was pretrained on by far the largest and most diverse corpus among the compared encoders (61,415~h from 92 datasets), and its authors report frozen representations that clearly outperform those of CBraMod, BIOT, and LaBraM~\citep{ouahidi2025reve}, whereas CBraMod and LaBraM are reported to require fine-tuning~\citep{wang2025cbramod,jiang2024labram}.
We therefore use the pretrained REVE for the subsequent experiments.
%

%
% --------------------------------
%
\subsection{Effect of decoder depth}
\label{sec:reve-initialization-depth}
%
% ------------------
%
\par
We vary the number $L$ of residual blocks in the decoder $\mathcal{M}_{\phi}$ defined in Sec.~\ref{subsec:eegencoder}, comparing frozen and fine-tuned REVE encoders in Tab.~\ref{tab:reve-initialization-depth}.
With the encoder frozen, $L=4$ gives the lowest errors, indicating that a strong decoder is needed for extracting the needed information from the frozen representation.
With fine-tuning, a more light-weight decoder is sufficient: $L=2$ gives the best metrics except for the LPIPS score.
Therefore, we use $L=2$ for the subsequent experiments.
%
%%%%
% Generated by paper/scripts/build_reve_initialization_depth.py; sources: data/reve_initialization_depth_sources.json.
\begin{table}[!t]
    \centering
    \caption{\textbf{REVE initialization and decoder-depth comparison.} REVE with $L=1$--$4$ decoder blocks on seen expression types held out in the test session, evaluated on Subject-03, comparing frozen and unfrozen EEG encoders.
%
    %All runs use one-second EEG and target windows.
    }
    \label{tab:reve-initialization-depth}
    \vspace{3pt}

    \begingroup
    \fontsize{8pt}{9.5pt}\selectfont
    \setlength{\tabcolsep}{3pt}
    \renewcommand{\arraystretch}{1.08}
    \begin{tabular*}{\textwidth}{@{\extracolsep{\fill}}lccccc@{}}
        \toprule
        \textbf{Decoder depth}
            & \textbf{LPIPS} $\downarrow$
            & \textbf{Blendshape MAE} $\downarrow$
            & \textbf{Latent MSE} $\downarrow$
            & \textbf{Mesh MAE [mm]} $\downarrow$
            & \textbf{Lmk Wing [mm]} $\downarrow$ \\
        \midrule
        \multicolumn{6}{l}{\textit{Pretrained frozen}} \\
        $L=1$ & 0.0650 & 0.0470 & 0.178 & 1.034 & 2.445 \\
        $L=2$ & 0.0643 & 0.0434 & 0.188 & 1.074 & 2.447 \\
        $L=3$ & 0.0624 & 0.0449 & \textbf{0.168} & 1.003 & 2.395 \\
        $L=4$ & \textbf{0.0608} & \textbf{0.0424} & \textbf{0.168} & \textbf{0.989} & \textbf{2.386} \\
        \midrule
        \multicolumn{6}{l}{\textit{Pretrained unfrozen}} \\
        $L=1$ & 0.0644 & 0.0465 & 0.197 & 1.054 & 2.525 \\
        $L=2$ & 0.0568 & \textbf{0.0345} & \textbf{0.164} & \textbf{0.943} & \textbf{2.195} \\
        $L=3$ & 0.0559 & 0.0385 & 0.188 & 1.033 & 2.436 \\
        $L=4$ & \textbf{0.0554} & 0.0389 & 0.206 & 1.068 & 2.424 \\
        \bottomrule
    \end{tabular*}
    \endgroup
\end{table}

%%%%
\FloatBarrier
%
% --------------------------------
%
\subsection{On the effectiveness of EEG pre-training}
\label{sec:results-subjects-splits}
In this experiment, we more systematically study the effects of unfreezing a pre-trained EEG encoder and show the results in \cref{tab:reve-initialization-subject-mean}.
First, as indicated above, we find that pre-training is beneficial, as from scratch training is substantially worse than both pre-trained variants.
Generally, unfreezing the encoder is effective for both, seen and unseen expressions, as the collected EEG data is still different to the pre-training data of REVE.
However, it can be observed that the performance gaps between the frozen and the unfrozen settings decrease for unseen expressions.
Hence, fine-tuning the encoder is less beneficial for out-of-distribution expressions.
This provides an interesting observation for future studies on the generalization capabilities of EEG foundational models.

%%%%
% Numerical sources: data/table_results_audit.json; blendshape MAE: data/reve_depth_august_matrix.json (subject_means, L=2); Pretrained unfrozen rows: the all-channel runs of Tab. 5 (data/table9_channel_masking_sources.json, six-subject means).
\begin{table}[!t]
    \centering
    \caption{\textbf{Effect of foundation-model initialization.} 
    Average results over all six subjects, evaluating held-out test sequences with expressions seen or unseen during training.}
    \label{tab:reve-initialization-subject-mean}
    \vspace{3pt}

    \begingroup
    \fontsize{8pt}{9.5pt}\selectfont % unscaled equivalent of the resized Tab. 2
    \setlength{\tabcolsep}{3pt}
    \renewcommand{\arraystretch}{1.08}
    \begin{tabular*}{\textwidth}{@{\extracolsep{\fill}}lccccc@{}}
        \toprule
        \textbf{Initialization}
            & \textbf{LPIPS} $\downarrow$
            & \textbf{Blendshape MAE} $\downarrow$
            & \textbf{Latent MSE} $\downarrow$
            & \textbf{Mesh MAE [mm]} $\downarrow$
            & \textbf{Lmk Wing [mm]} $\downarrow$ \\
        \midrule
        \multicolumn{6}{l}{\textit{Seen expressions (S; held-out session for each expression)}} \\
        Scratch & 0.0736 & 0.0463 & 0.270 & 1.237 & 2.820 \\
        Pretrained frozen & 0.0701 & 0.0453 & 0.235 & 1.163 & 2.611 \\
        Pretrained unfrozen & \textbf{0.0651} & \textbf{0.0412} & \textbf{0.214} & \textbf{1.046} & \textbf{2.430} \\
        \midrule
        \multicolumn{6}{l}{\textit{Unseen expressions (U; absent from training)}} \\
        Scratch & 0.0791 & 0.0581 & 0.305 & 1.305 & 2.897 \\
        Pretrained frozen & 0.0735 & 0.0532 & 0.262 & 1.182 & 2.652 \\
        Pretrained unfrozen & \textbf{0.0687} & \textbf{0.0487} & \textbf{0.246} & \textbf{1.134} & \textbf{2.508} \\
        \bottomrule
    \end{tabular*}
    \endgroup
\end{table}

%%%%
\FloatBarrier

%
% ------------------
%
%
\subsection{On the Informativeness of Different Brain Areals for Face Decoding} \label{sec:area}
%
%%%%
% Generated by paper/scripts/build_table9_channel_masking.py; sources: data/table9_channel_masking_sources.json.
\begin{table}[!t]
    \centering
    \caption{\textbf{Channel-masking ablations.} 
    Average subject results using pretrained-unfrozen REVE-base and the $L=2$ decoder.
    Numbers in parentheses give the number of zeroed channels; Sec.~\ref{subsec:implementation} lists the electrodes.
}
    \label{tab:ablation-channel-masking}
    \vspace{3pt}

    \begingroup
    \fontsize{8pt}{9.5pt}\selectfont
    \setlength{\tabcolsep}{3pt}
    \renewcommand{\arraystretch}{1.08}
    \begin{tabular*}{\textwidth}{@{\extracolsep{\fill}}>{\raggedright\arraybackslash}p{0.2\linewidth}ccccc@{}}
        \toprule
        \textbf{Variant (zeroed)}
            & \textbf{LPIPS} $\downarrow$
            & \textbf{Blendshape MAE} $\downarrow$
            & \textbf{Latent MSE} $\downarrow$
            & \textbf{Mesh MAE [mm]} $\downarrow$
            & \textbf{Lmk Wing [mm]} $\downarrow$ \\
        \midrule
        \multicolumn{6}{l}{\textit{Seen expressions (S; held-out session for each expression)}} \\
        All channels (0) & 0.0651 & 0.0412 & \textbf{0.214} & 1.046 & 2.430 \\
        Bad channels (4--19) & 0.0668 & \textbf{0.0398} & 0.219 & \textbf{1.039} & \textbf{2.426} \\
        Motor region (20) & \textbf{0.0649} & 0.0408 & 0.216 & 1.051 & 2.439 \\
        Frontal (15) & 0.0676 & 0.0433 & 0.221 & 1.082 & 2.537 \\
        Occipital (10) & 0.0669 & 0.0427 & 0.221 & 1.077 & 2.486 \\
        Temporal (10) & 0.0674 & 0.0439 & 0.234 & 1.100 & 2.519 \\
        \midrule
        \multicolumn{6}{l}{\textit{Unseen expressions (U; absent from training)}} \\
        All channels (0) & 0.0687 & 0.0487 & 0.246 & 1.134 & 2.508 \\
        Bad channels (4--19) & 0.0694 & 0.0485 & 0.249 & 1.139 & 2.522 \\
        Motor region (20) & 0.0701 & 0.0492 & 0.252 & 1.145 & 2.536 \\
        Frontal (15) & \textbf{0.0682} & \textbf{0.0482} & \textbf{0.241} & \textbf{1.122} & \textbf{2.505} \\
        Occipital (10) & 0.0703 & 0.0489 & 0.252 & 1.160 & 2.547 \\
        Temporal (10) & 0.0694 & 0.0492 & 0.255 & 1.153 & 2.550 \\
        \bottomrule
    \end{tabular*}
    \endgroup
\end{table}

%%%%

To analyze what the model has learned, we study the effect of removing electrodes corresponding to various brain areas, similar to region-wise electrode ablations in prior EEG decoding work~\citep{song2024decoding,hu2026see}.
For this, we retrain models where the respective channels are masked out, i.e., set to zero.
Masking motor or prefrontal channels degrades facial expressions reconstructed from EEG~\citep{hu2026see}.
Furthermore, the frontal and temporal groups also lie closest to the frontalis and temporalis muscles, whose EMG is strongest at these sites and attenuates toward central and posterior sites~\citep{goncharova2003emg}, whereas the occipital group lies over the visual cortex~\citep{song2024decoding}.
Further, we also study the effect of masking out bad channels, which \cite{bigdelyshamlo2015prep} identify by extreme amplitudes, high-frequency noise, or low correlation with, and poor predictability from, the other channels.
Bad channels are detected per subject on training expressions with PyPREP, a Python implementation of the PREP pipeline~\cite{bigdelyshamlo2015prep}.
The results are presented in \cref{tab:ablation-channel-masking}.
The findings reveal that for seen in-distribution expressions, a substantial amount of the performance can be attributed to the temporal electrodes, as removing the influence of the those electrodes results in a substantial performance drop. 
Performance drops are observed when masking other brain areas, including  the motor cortex.
In contrast, removing bad channels reduces the effect of artifacts in the EEG, eventually even improving the performance compared to experiment where access to all electrodes is given.
Interestingly, for unseen expressions, the effect of removing electrodes capturing the frontal brain areal is contrary compared to the seen expression: 
Results substantially improve when their effect is removed.
We hypothesize that this could be attributed to the fact that the motion artifacts in the frontal brain area observed in the EEG signal provide less consistent information for unseen expressions than for seen expressions.
In contrast, the signals obtained from the motor cortex electrodes are more consistent, resulting in a performance drops that are comparable to masking temporal brain areas.

%
% --------------------------------
%
\subsection{Qualitative Results}
\label{sec:qualitative-seen-unseen}
%
%%%%
\begin{figure}[t]
    \centering
%%%%
\begingroup
\setlength{\tabcolsep}{0pt}
\renewcommand{\arraystretch}{1}
\begin{tabular}{@{}c@{\hspace{3pt}}c@{\hspace{2pt}}c@{\hspace{8pt}}c@{\hspace{2pt}}c@{}}
& \multicolumn{2}{c@{\hspace{8pt}}}{\small Seen expression types} & \multicolumn{2}{c}{\small Unseen expression types} \\[2pt]
& \small Ours & \small Ground truth & \small Ours & \small Ground truth \\[3pt]
\raisebox{0.065\linewidth}{\rotatebox{90}{\scriptsize Subject-01}} & \includegraphics[width=0.232\linewidth]{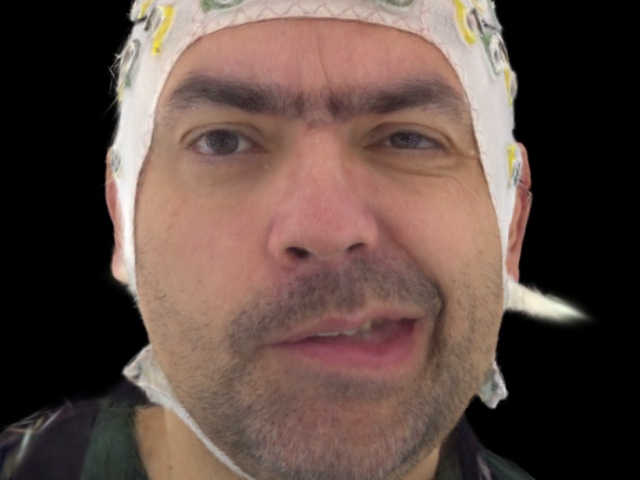} & \includegraphics[width=0.232\linewidth]{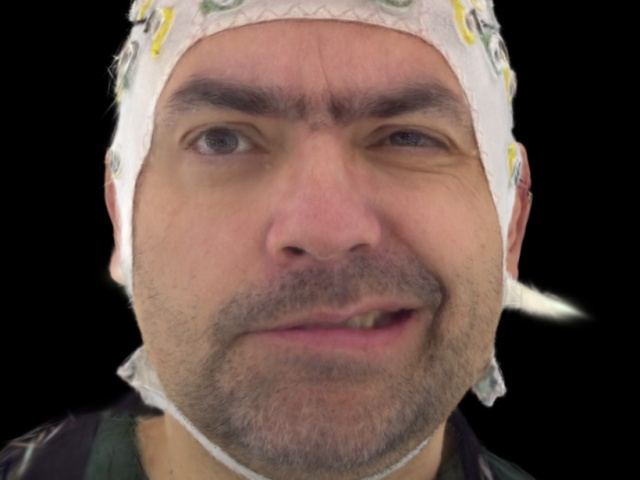} & \includegraphics[width=0.232\linewidth]{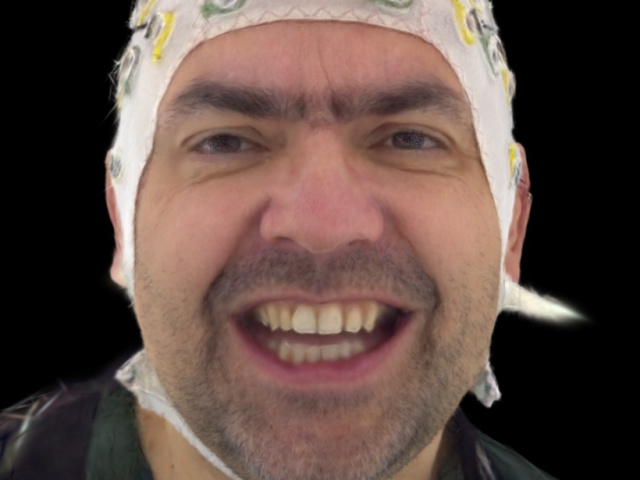} & \includegraphics[width=0.232\linewidth]{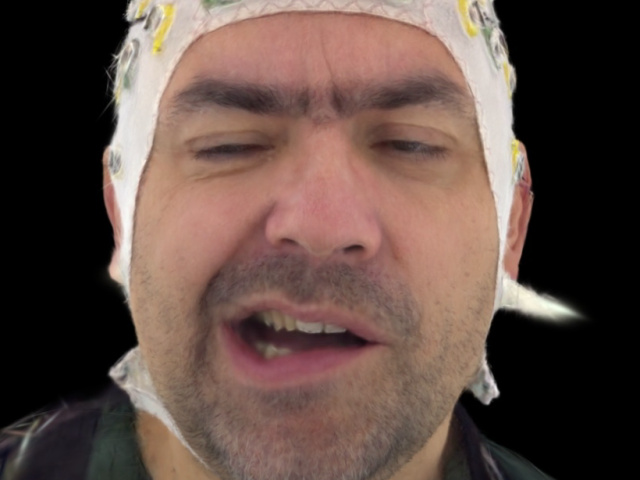} \\[-2pt]
& \multicolumn{2}{c@{\hspace{8pt}}}{\small\itshape mouth and nose left} & \multicolumn{2}{c}{\small\itshape jaw right with teeth} \\[3pt]
\raisebox{0.065\linewidth}{\rotatebox{90}{\scriptsize Subject-04}} & \includegraphics[width=0.232\linewidth]{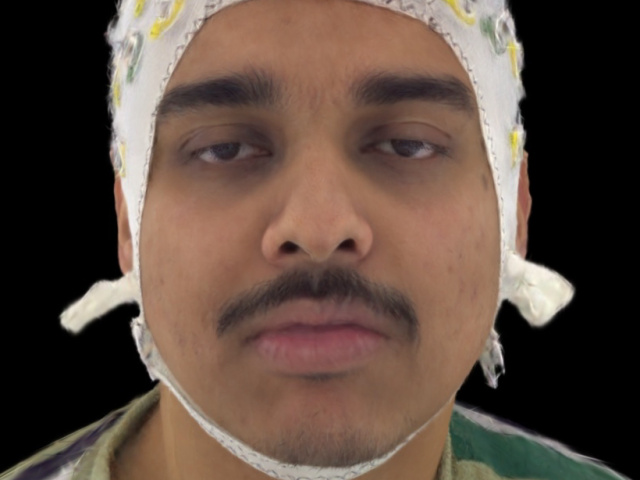} & \includegraphics[width=0.232\linewidth]{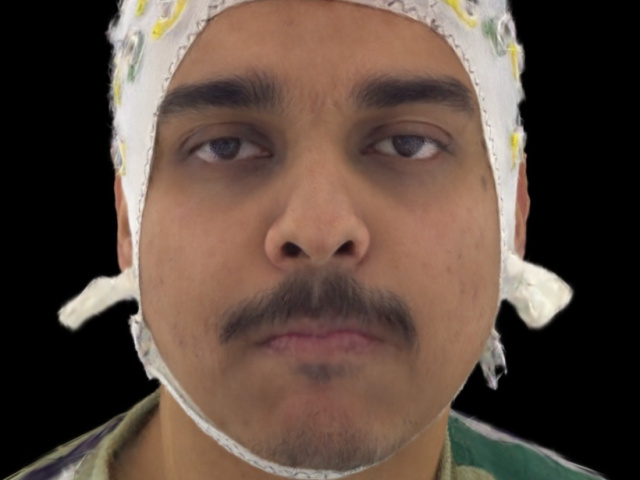} & \includegraphics[width=0.232\linewidth]{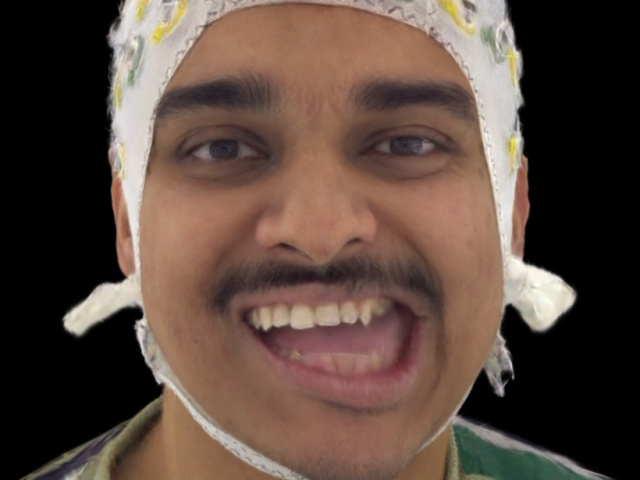} & \includegraphics[width=0.232\linewidth]{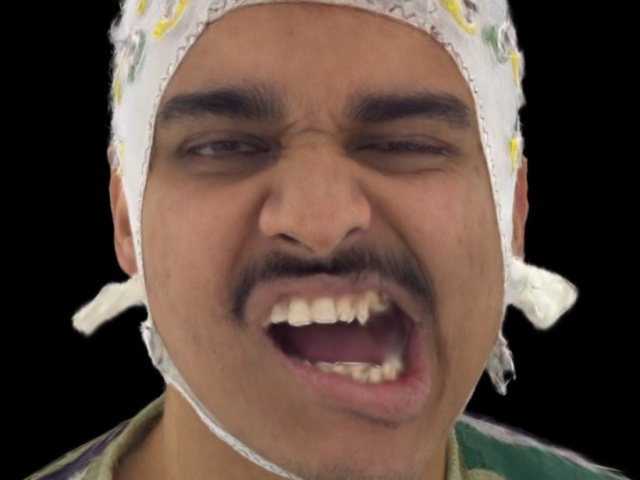} \\[-2pt]
& \multicolumn{2}{c@{\hspace{8pt}}}{\small\itshape left cheek puffed} & \multicolumn{2}{c}{\small\itshape jaw left with teeth} \\[3pt]
\raisebox{0.065\linewidth}{\rotatebox{90}{\scriptsize Subject-05}} & \includegraphics[width=0.232\linewidth]{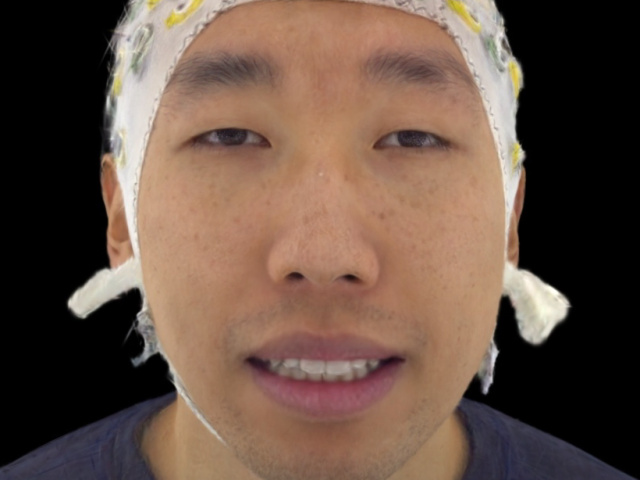} & \includegraphics[width=0.232\linewidth]{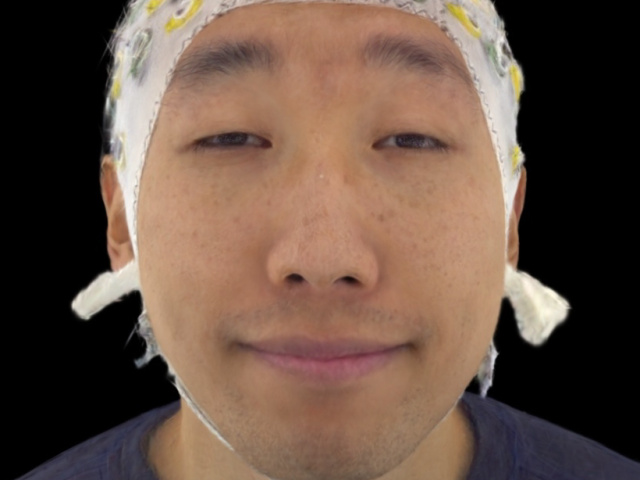} & \includegraphics[width=0.232\linewidth]{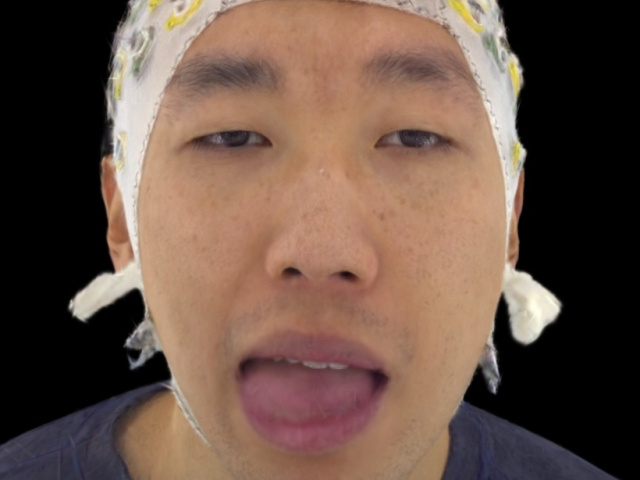} & \includegraphics[width=0.232\linewidth]{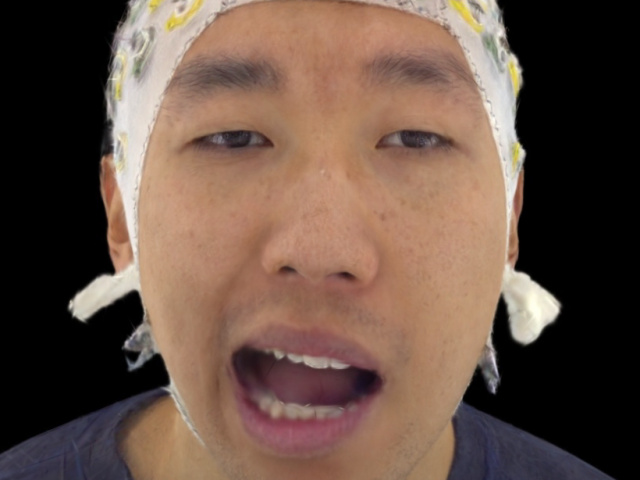} \\[-2pt]
& \multicolumn{2}{c@{\hspace{8pt}}}{\small\itshape stretched smile} & \multicolumn{2}{c}{\small\itshape jaw right with teeth} \\[3pt]
\end{tabular}
\endgroup

%%%%
    \caption{\textbf{Qualitative reconstruction of seen and unseen expression types.}
    Each row shows one subject, with Ours and Ground truth paired for each expression group.
    Ground truth denotes the fitted Gaussian-avatar render at the matching frame.
%
    %Predictions use the pretrained-unfrozen model; snapshots are taken at approximately $t=3$~s.
%
    %All seen examples are from session 2; unseen examples are from session 2 for Subjects 01 and 04 and session 1 for Subject-05.
    }
    \label{fig:qualitative-seen-unseen}
\end{figure}

%%%%
% ------------------
%
\par
Fig.~\ref{fig:qualitative-seen-unseen} compares EEG-driven predictions with fitted-avatar ground truth for three subjects using the pretrained-unfrozen model.
The examples show recognizable expression structure in both groups: the seen mouth-and-nose asymmetry of Subject-01 and the unseen open-mouth expressions of Subjects 04 and 05 are reflected in the predictions.
Fine details remain imperfect, including lip closure in Subject-05's seen example and the mouth shape and eye closure in Subject-01's unseen example.
As each pair shares the same personalized avatar, the comparison shows expression reconstruction rather than identity synthesis.
%
%
% --------------------------------
%

%%%%
%
% ================================
%
\section{Limitations and Future Work}
\label{sec:futurework}
%
% ------------------
%
\par
Our initial dataset contains only six participants recorded in two sessions each, motivating larger studies with more participants and recording sessions to assess generalization.
As recorded EEG signal are substantially noisier than classical image-driven signals, accurate facial tracking is more challenging with current reconstructions not capturing the level of detail available in image-driven approaches.
Future work could explore further analyses with complementary sensing modalities, such as functional near-infrared spectroscopy (fNIRS), to investigate whether they provide additional information for facial expression decoding.
Finally, our framework models only facial expressions and does not account for movements of other body parts - movements that could also impact the signal quality.
Future work can also analyze further which EEG signal components drive the prediction of which facial articulations.

%%%%
%
%%%%%%%%%%%%%%%%%%%%%%%%%%%%%%%%%
%
\section{Conclusion} \label{sec:conclusion}
In this work, we introduced the task of mapping EEG signals to face articulations.
For this, we presented a systematically captured dataset that allows studying the relation between the two modalities.
Subjects were recorded with an EEG and captured in a multi-camera rig that allows highly accurate reconstructions.
We further presented a surprisingly effective first baseline to predict face articulations and we studied various design dimensions including the effect of using pre-trained EEG foundational models, the importance of using expressive decoders, and the importance of different brain areals for the prediction of facial expressions.
We hope that the dataset and the first analysis serves the community and encourages further analysis on the informativeness of EEG signals to infer facial geometry and articulations.
%
%%%%%%%%%%%%%%%%%%%%%%%%%%%%%%%%%
%

\section*{AI disclosure}

We used AI tools to refine text formatting, grammar, and language, assist with table preparation, and support code development and figure generation through agentic coding. AI-assisted outputs were reviewed and verified by the authors.

\section*{Ethics statement}

Facial avatars and EEG data were handled with respect for participant privacy, consent, and data security. Their use was limited to the agreed research purposes. Facial avatars were not used to impersonate individuals or misrepresent their identity.

%%%%

\bibliographystyle{plainnat_no_doi}
\bibliography{main}

\clearpage

\appendix
%%%%
%
% ================================
%

 {
        \centering
        \Large
        Supplementary Material \\
        \vspace{1.5em}
   }

\section{More Dataset and Method Details}
\label{app:method-details}
%
% ------------------
%
\par
This supplement describes synchronization (Sec.~\ref{subsec:sync}), dataset statistics and splits (Sec.~\ref{app:dataset-splits}), avatar fitting and inversion (Sec.~\ref{app:avatar-fitting}), training objectives (Sec.~\ref{subsec:training_objective}), and implementation details (Sec.~\ref{subsec:implementation}).
The expression repertoire is listed in Sec.~\ref{app:expression-repertoire}.

%
%%%%
\suppressfloats[t]
% Brain2Face overview figure.
%%%%%%%%%%%%%%%%%%%%%%%%%%%%%%%%%%%%%%
%
\begin{figure*}[t]
\begingroup
\centering
\makebox[\textwidth][c]{%
\begin{minipage}{\textwidth}
\centering
\includegraphics[width=\linewidth]{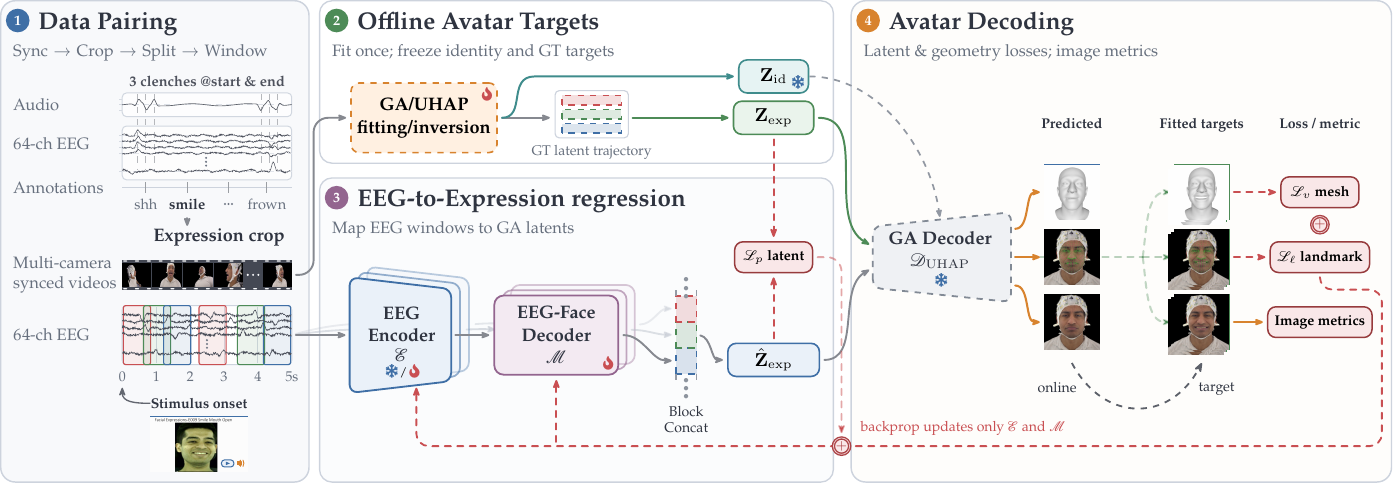}
\caption[\BrainTwoFace{} overview]{%
\textbf{\BrainTwoFace{} overview.}
A synchronized recording segment aligns annotated multi-view video with preprocessed (Sec.~\ref{subsec:eegencoder}) EEG. Expression groups are assigned to training and test sets
(Tab.~\ref{tab:dataset-splits}) before windows are sampled. Offline UHAP avatar fitting
produces the subject identity code \(\mathbf{Z}_{\mathrm{id}}\); a frozen image encoder
produces the expression trajectory $\mathbf{Z}_{\mathrm{exp}}$, which is the direct
regression target for the EEG encoder \(\mathcal{E}\) (frozen or fine-tuned) and decoder \(\mathcal{M}\). Consecutive EEG windows are processed with shared
weights, and their predicted chunks are concatenated into a continuous latent
sequence \(\hat{\mathbf{Z}}_{\mathrm{exp}}\), which drives the frozen UHAP decoder
\(\mathcal{D}_{\mathrm{UHAP}}\). Latent loss \(\mathcal{L}_p\) compares the
prediction against the target expression codes. Mesh and landmark losses
\(\mathcal{L}_v\) and \(\mathcal{L}_{\ell}\) supervise decoded geometry and
backpropagate through the frozen avatar decoder, the EEG-to-expression decoder \(\mathcal{M}\) and, when unfrozen, the EEG encoder are updated.
The auxiliary loss \(\mathcal{L}_{\mathrm{aux}}\) (Sec.~\ref{subsec:supervision}) is omitted for clarity.
%Image metrics use matched predicted and fitted ground-truth renders for evaluation and do not contribute gradients.
}
\label{fig:overview}
\end{minipage}%
}%
\endgroup
\end{figure*}
%
%%%%%%%%%%%%%%%%%%%%%%%%%%%%%%%%%%%%%%
%

%%%%
%
% --------------------------------
%
\subsection{System Synchronization} \label{subsec:sync}
%
% ------------------
%
\par
To synchronize the EEG signals with the facial video recordings, a synchronization event is manually triggered for both the face and EEG recording devices.
Additionally, jaw clenching serves as a complementary synchronization signal, as it produces large-amplitude artifacts that are easily detectable visually in the video, in its audio track, and in the EEG traces.
These events are used to compute a time-lag offset, enabling precise temporal alignment of the two modalities.
%
% --------------------------------
%
\subsection{Dataset statistics and splits}
\label{app:dataset-splits}
%
% ------------------
%
\par
Brain2Face contains recordings from six adult male subjects aged 24 to 47, each recorded in two synchronized EEG and video sessions.
The 12 sessions total approximately 4~h 15~min of capture time, excluding the resting EEG baselines.
Cameras recorded at $3840\times2160$ resolution; we downsample the images to $1920\times1080$ for avatar fitting and inversion.
Fig.~\ref{fig:dataset} illustrates expression variation within a subject, simultaneous views of the same expression, and appearance variation across subjects.
%
% ------------------
%
\par
The protocol covers 75 predefined expression types per session.
For the REVE experiments, we extract five-second paired segments after the annotated expression onsets.
The processed dataset retains 891 such segments across the 12 sessions, totaling 74~min 15~s of expression data.
Each segment pairs 64-channel EEG at 200~Hz with 256-dimensional avatar expression codes at 29.97~fps.
%
% ------------------
%
\par
Four fixed, disjoint expression groups contain 45 types in $b_1$ and 10 each in $b_2$, $b_3$, and $b_4$.
Sec.~\ref{app:expression-repertoire} lists all expression names and their group membership.
Seen (S) and unseen (U) expressions are defined in Sec.~\ref{sec:evaluation-protocol}.
Tab.~\ref{tab:dataset-splits} sums the retained segments across the six subject datasets using these categories.
%%%%
%%%%
\begin{table}[!htbp]
    \centering
    \caption{\textbf{Cross-session split.}
    Cells list the expression groups assigned to each role in each session.
    Counts give retained five-second segments and their percentage of the 891 segments across six subjects.
    Each subject has a separate model.
    }
    \label{tab:dataset-splits}
    \begingroup
    \small
    \setlength{\tabcolsep}{8pt}
    \renewcommand{\arraystretch}{1.08}
    \begin{tabular}{@{}lccc@{}}
        \toprule
        & & \multicolumn{2}{c}{Test sequences} \\
        \cmidrule(lr){3-4}
        & Training (T) & \shortstack{Seen expressions\\(S)} & \shortstack{Unseen expressions\\(U)} \\
        \midrule
        Session~1 & $b_1, b_2$ & $b_3$ & $b_4$ \\
        Session~2 & $b_1, b_3$ & $b_2$ & $b_4$ \\
        Segments (\%) & 655 (73.5) & 119 (13.4) & 117 (13.1) \\
        \bottomrule
    \end{tabular}
    \endgroup
\end{table}

%%%%
%
% --------------------------------
%
%
%%%%
%%%%

%
% --------------------------------
%
\subsection{Gaussian avatar fitting and inversion}
\label{app:avatar-fitting}
%
% ------------------
%
\par

Fig.~\ref{fig:ga-fitting} summarizes how calibrated multi-view expression frames provide RGB images, foreground masks, and triangulated 3D guide points.
A frozen gaze-conditioned image encoder from UHAP~\cite{teotia2025audiodrivenuniversalgaussianhead} produces the per-frame expression codes $\mathbf{Z}_{\mathrm{exp}}^t$.
Fitting adapts the participant's identity code and decoder weights, together with pose and color calibration, using the multi-view observations and 3D guides.
Masked RGB and mouth losses compare rendered and observed colors, while a guide loss matches 200 selected 3D points.
Structural similarity and VGG-based perceptual losses compare image appearance.
Regularization constrains the identity code and Gaussian scales, offsets, and opacities; a Laplacian loss encourages guide-mesh smoothness, and a locality loss penalizes deviations from the pretrained decoder.
After fitting and guide-mesh refinement, the identity code and personalized decoder are frozen, and the saved expression trajectory serves as the EEG regression target.
%
% ------------------
%
\par
The decoder supplies a guide mesh with $N_v=7306$ vertices in millimetres.
The fixed selection operator $\Pi$ extracts the $K_\ell=200$ guide-mesh landmarks that the fitting guide loss also uses from both the fitted and predicted meshes.
Thus, the landmark loss below compares 3D coordinates, rather than image-space projections.
The same frozen decoder and camera settings produce the predicted and fitted-avatar renders; the 52-dimensional blendshape vectors are used only for evaluation.
%
%%%%
\input{fig/ga_fitting/ga_fitting}
%%%%

%
% --------------------------------
%
\subsection{Training Objective} \label{subsec:training_objective}
%
% ------------------
%
\par
The combined objective as introduced in Eq.~\eqref{eq:total_loss} in the main paper uses latent, mesh, landmark, and auxiliary trajectory losses with $(\lambda_p,\lambda_v,\lambda_\ell,\lambda_{\mathrm{aux}})=(1,5,10,0.5)$.
%
% ------------------
%
\par
For a batch of $B$ windows, $\mathbf{Z}_{\mathrm{exp},b}^{t}$ denotes the length-$d$ vector at frame $t$ of window $b$.
The sets $\Omega_v,\Omega_\ell\subseteq\{1,\dots,B\}\times\{1,\dots,N_z\}$ contain the window--frame pairs $(b,t)$ at which the mesh and landmark losses are evaluated.
The expression-latent loss averages squared error over the batch:
%%%%
\begin{equation}
    \mathcal{L}_p(\hat{\mathbf{Z}}_{\mathrm{exp}},\mathbf{Z}_{\mathrm{exp}}) = \frac{1}{BdN_z}\sum_{b=1}^{B}\sum_{t=1}^{N_z}
    \left\|\hat{\mathbf{Z}}_{\mathrm{exp},b}^{t}-\mathbf{Z}_{\mathrm{exp},b}^{t}\right\|_2^2.
\end{equation}
%%%%
The mesh loss is the mean absolute coordinate error:
%%%%
\begin{equation}
    \mathcal{L}_v(\hat{\mathbf{V}},\mathbf{V}) = \frac{1}{3N_v|\Omega_v|}
    \sum_{(b,t)\in\Omega_v}\sum_{v=1}^{N_v}\sum_{a=1}^{3}
    \left|\hat V^{t}_{b,v,a}-V^{t}_{b,v,a}\right|.
\end{equation}
%%%%
The landmark loss applies a coordinate-wise Wing penalty:
%%%%
\begin{equation}
    \mathcal{L}_\ell(\hat{\mathbf{L}},\mathbf{L}) = \frac{1}{3K_\ell|\Omega_\ell|}
    \sum_{(b,t)\in\Omega_\ell}\sum_{k=1}^{K_\ell}\sum_{a=1}^{3}
    \rho_{\mathrm{W}}(\hat L^{t}_{b,k,a}-L^{t}_{b,k,a}),
\end{equation}
%%%%
where, for $u=|e|$,
%%%%
\begin{equation}
    \rho_{\mathrm{W}}(e)=
    \begin{cases}
        w_{\mathrm{W}}\log(1+u/\epsilon_{\mathrm{W}}),&u<w_{\mathrm{W}},\\
        u-C_{\mathrm{W}},&u\ge w_{\mathrm{W}},
    \end{cases}
    \quad
    C_{\mathrm{W}}=w_{\mathrm{W}}-w_{\mathrm{W}}\log(1+w_{\mathrm{W}}/\epsilon_{\mathrm{W}}).
\end{equation}
%%%%
We use the Wing form~\cite{feng2018wing} with $w_{\mathrm{W}}=10$~mm and $\epsilon_{\mathrm{W}}=2$~mm.
Both $\mathcal{L}_v$ and $\mathcal{L}_\ell$ have units of mm.
%
% ------------------
%
\par
The auxiliary loss applies the latent loss to the auxiliary sequence, $\mathcal{L}_{\mathrm{aux}}=\mathcal{L}_p(\hat{\mathbf{Z}}_{\mathrm{exp}}^{\mathrm{aux}},\mathbf{Z}_{\mathrm{exp}})$.
%

%
% --------------------------------
%
\subsection{Implementation Details} \label{subsec:implementation}
%
% ------------------
%
\par
With one-second windows, REVE returns one token per electrode, so the decoder's kernel-size-3 residual blocks, with increasing dilation, operate across the 64 electrode tokens.
Two strided depthwise-separable convolutions reduce these tokens to 16, linear interpolation resamples them to $N_z=30$ frames, and a framewise MLP maps each frame to $d=256$ coordinates.
The MLP decoder used for encoder probing flattens the encoder features and applies two 512-unit hidden layers with GEGLU, RMSNorm, and dropout probability 0.75 before projecting to $256N_z$ values; it has no auxiliary head.
%
% ------------------
%
\par
EEG models use AdamW under three initialization regimes: pretrained-frozen, pretrained-unfrozen, and scratch.
The first keeps the pretrained encoder fixed, the second fine-tunes it, and the third trains it from random initialization.
%
% ------------------
%
\par
During training, windows are sampled at random positions within the training segments.
%
% ------------------
%
\par
All results use predictions from consecutive non-overlapping windows, whereas the qualitative clips concatenate them.
%
% ------------------
%
\par
The channel-masking ablation (Tab.~\ref{tab:ablation-channel-masking}) zeroes the following electrodes: motor, FC1--FC6, C1--C6, Cz, CP1--CP6, and CPz; frontal, Fp1, Fp2, AF3, AF4, AF7, AF8, F1--F8, and Fz; occipital, O1, O2, Oz, PO3, PO4, PO7--PO10, and POz; temporal, T7, T8, FT7--FT10, and TP7--TP10.

%%%%
%%%%
%
% ================================
%
\section{List of Expressions}
\label{app:expression-repertoire}

Tab.~\ref{tab:expression-repertoire} lists all 75 expression names where expressions from the group $b_1$ are training expression in session 1 and 2, $b_2$ are training expressions for session 1, $b_3$ are training expressions for session 2, and $b_4$ are only used for evaluation from both sessions.
%The labels match the dataset annotations. The expression \texttt{jaw\_clench} in $b_3$ is distinct from the repeated jaw clenches used for synchronization (Sec.~\ref{subsec:sync}).

%%%%
\begingroup
\small
\setlength{\LTcapwidth}{\textwidth}
\renewcommand{\arraystretch}{1.02}
\begin{longtable}{@{}rcl@{}}
\caption{\textbf{Full expression repertoire.}
The 75 expression labels and their fixed group assignments. Indices follow
this group order and do not indicate recording order.}
\label{tab:expression-repertoire}\\
\toprule
Index & Group & Expression name \\
\midrule
\endfirsthead
\multicolumn{3}{@{}l}{\tablename~\thetable\ continued}\\
\toprule
Index & Group & Expression name \\
\midrule
\endhead
\midrule
\multicolumn{3}{r@{}}{\small Continued on the next page}\\
\endfoot
\bottomrule
\endlastfoot
1 & $b_1$ & \texttt{lips\_puffed} \\
2 & $b_1$ & \texttt{raise\_left\_eyebrow} \\
3 & $b_1$ & \texttt{relaxed\_mouth\_open} \\
4 & $b_1$ & \texttt{left\_side\_sneer} \\
5 & $b_1$ & \texttt{show\_all\_teeth} \\
6 & $b_1$ & \texttt{jaw\_open\_pull\_lips\_in} \\
7 & $b_1$ & \texttt{kiss\_lips\_look\_down} \\
8 & $b_1$ & \texttt{stick\_lower\_lip\_out} \\
9 & $b_1$ & \texttt{mouth\_open\_tongue\_out} \\
10 & $b_1$ & \texttt{smile\_mouth\_closed} \\
11 & $b_1$ & \texttt{jaw\_open\_sharp\_corner\_lip\_stretch} \\
12 & $b_1$ & \texttt{suck\_cheeks\_in\_brows\_up} \\
13 & $b_1$ & \texttt{neck\_stretch\_brows\_up} \\
14 & $b_1$ & \texttt{neutral\_eyes\_open} \\
15 & $b_1$ & \texttt{nostrils\_sucked\_in} \\
16 & $b_1$ & \texttt{eyes\_wide\_open} \\
17 & $b_1$ & \texttt{bite\_lower\_lip} \\
18 & $b_1$ & \texttt{lips\_together\_pushed\_forward} \\
19 & $b_1$ & \texttt{tongue\_out\_flat} \\
20 & $b_1$ & \texttt{open\_lips\_mouth\_stretch\_nose\_wrinkled} \\
21 & $b_1$ & \texttt{tongue\_out\_left\_teeth\_showing} \\
22 & $b_1$ & \texttt{tongue\_over\_upper\_lip} \\
23 & $b_1$ & \texttt{jaw\_back} \\
24 & $b_1$ & \texttt{suck\_cheeks\_in} \\
25 & $b_1$ & \texttt{jaw\_open\_lips\_together} \\
26 & $b_1$ & \texttt{blow\_cheeks\_full\_of\_air} \\
27 & $b_1$ & \texttt{tongue\_out\_lips\_closed} \\
28 & $b_1$ & \texttt{lower\_eyebrows} \\
29 & $b_1$ & \texttt{hide\_lips\_look\_up} \\
30 & $b_1$ & \texttt{jaw\_open\_upper\_lip\_raised} \\
31 & $b_1$ & \texttt{raise\_left\_eyebrow\_pucker\_lips} \\
32 & $b_1$ & \texttt{tongue\_out\_rolled} \\
33 & $b_1$ & \texttt{frown\_stick\_lower\_lip\_out} \\
34 & $b_1$ & \texttt{wink\_left\_eye\_smile} \\
35 & $b_1$ & \texttt{frown} \\
36 & $b_1$ & \texttt{cheeks\_puffed} \\
37 & $b_1$ & \texttt{jaw\_open\_huge\_smile} \\
38 & $b_1$ & \texttt{raise\_right\_eyebrow} \\
39 & $b_1$ & \texttt{shh} \\
40 & $b_1$ & \texttt{jaw\_forward} \\
41 & $b_1$ & \texttt{pressed\_lips\_brows\_down} \\
42 & $b_1$ & \texttt{raise\_both\_eyebrows\_pucker\_lips} \\
43 & $b_1$ & \texttt{open\_mouth\_stretch\_nose\_wrinkled} \\
44 & $b_1$ & \texttt{bite\_tongue} \\
45 & $b_1$ & \texttt{tongue\_out\_right\_teeth\_showing} \\
\addlinespace
46 & $b_2$ & \texttt{wink\_right\_eye\_smile} \\
47 & $b_2$ & \texttt{right\_side\_sneer} \\
48 & $b_2$ & \texttt{right\_cheek\_puffed} \\
49 & $b_2$ & \texttt{left\_cheek\_puffed} \\
50 & $b_2$ & \texttt{mouth\_nose\_left} \\
51 & $b_2$ & \texttt{jaw\_open\_lips\_pushed\_out} \\
52 & $b_2$ & \texttt{swallow} \\
53 & $b_2$ & \texttt{scrunch\_face\_squeeze\_eyes} \\
54 & $b_2$ & \texttt{smile\_stretched} \\
55 & $b_2$ & \texttt{mouth\_nose\_right} \\
\addlinespace
56 & $b_3$ & \texttt{nostrils\_dilated} \\
57 & $b_3$ & \texttt{open\_mouth\_wide\_tongue\_up\_and\_back} \\
58 & $b_3$ & \texttt{raise\_inner\_eyebrows} \\
59 & $b_3$ & \texttt{suck\_cheeks\_in\_brows\_frowned} \\
60 & $b_3$ & \texttt{oooo} \\
61 & $b_3$ & \texttt{jaw\_clench} \\
62 & $b_3$ & \texttt{raise\_upper\_lip\_scrunch\_nose} \\
63 & $b_3$ & \texttt{raise\_right\_eyebrow\_pucker\_lips} \\
64 & $b_3$ & \texttt{scream\_eyebrows\_up} \\
65 & $b_3$ & \texttt{tongue\_out\_thick} \\
\addlinespace
66 & $b_4$ & \texttt{neutral\_eyes\_closed} \\
67 & $b_4$ & \texttt{lips\_open\_right} \\
68 & $b_4$ & \texttt{jaw\_drop\_brows\_up} \\
69 & $b_4$ & \texttt{bite\_upper\_lip} \\
70 & $b_4$ & \texttt{jaw\_open\_mouth\_corners\_down\_nose\_wrinkled} \\
71 & $b_4$ & \texttt{raise\_cheeks} \\
72 & $b_4$ & \texttt{mouth\_open\_jaw\_right\_show\_teeth} \\
73 & $b_4$ & \texttt{lips\_open\_left} \\
74 & $b_4$ & \texttt{smile\_mouth\_open} \\
75 & $b_4$ & \texttt{mouth\_open\_jaw\_left\_show\_teeth} \\
\end{longtable}
\endgroup

%%%%

%
% ================================
%
% %
% %
% %
% % ================================
% %
% %
% %
% %

%%%%
%%%%
%%%%
%%%%
%%%%

\end{document}